\documentclass{article}

     \usepackage[preprint]{neurips_2019}

\usepackage[utf8]{inputenc} 
\usepackage[T1]{fontenc}    
\usepackage{hyperref}       
\usepackage{url}            
\usepackage{booktabs}       
\usepackage{amsfonts}       
\usepackage{nicefrac}       
\usepackage{microtype}      
\usepackage{overpic}
\usepackage{graphicx}
\usepackage{subcaption}
\usepackage{caption}
\usepackage{color}
\usepackage{amsmath}
\usepackage{siunitx}

\usepackage{varwidth}
\DeclareCaptionFormat{myformat}{%
  \begin{varwidth}{\linewidth}%
    \centering
    #1#2#3%
  \end{varwidth}%
}

\newcommand{\R}{\mathbb{R}}

\newcommand{\norm}[1]{\left | \left | #1 \right | \right |}

\title{Deep Model Predictive Control with Online Learning for Complex Physical Systems}

\author{%
  Katharina Bieker\thanks{Chair of Applied Mathematics, Paderborn University, Germany}\\
  \texttt{bieker@math.upb.de} \\
  \And
  Sebastian Peitz\footnotemark[1] \\
  \texttt{speitz@math.upb.de} \\
  \AND
  Steven L. Brunton\thanks{Department of Mechanical Engineering, University of Washington, Seattle, WA 98195}\\
  \texttt{sbrunton@uw.edu} \\
  \And
  J. Nathan Kutz\thanks{Department of Applied Mathematics, University of Washington, Seattle, WA 98195}\\
  \texttt{kutz@uw.edu} \\
  \And
  Michael Dellnitz\footnotemark[1]\\
  \texttt{dellnitz@math.upb.de} \\
}

\begin{document}

\maketitle

\begin{abstract}
The control of complex systems is of critical importance in many branches of science, engineering, and industry. 
Controlling an unsteady fluid flow is particularly important, as flow control is a key enabler for technologies in energy (e.g., wind, tidal, and combustion), transportation (e.g., planes, trains, and automobiles), security (e.g., tracking airborne contamination), and health (e.g., artificial hearts and artificial respiration). 
However, the high-dimensional, nonlinear, and multi-scale dynamics make real-time feedback control infeasible. 
Fortunately, these high-dimensional systems exhibit dominant, low-dimensional patterns of activity that can be exploited for effective control in the sense that knowledge of the entire state of a system is not required. Advances in machine learning have the potential to revolutionize flow control given its ability to extract principled, low-rank feature spaces characterizing such complex systems. 
We present a novel deep learning model predictive control (DeepMPC) framework that exploits low-rank features of the flow in order to achieve considerable improvements to control performance. 
Instead of predicting the entire fluid state, we use a recurrent neural network (RNN) to accurately predict the control relevant quantities of the system. The RNN is then embedded into a MPC framework to construct a feedback loop, and incoming sensor data is used to perform online updates to improve prediction accuracy.
The results are validated using varying fluid flow examples of increasing complexity.

\end{abstract}

\section{Introduction}
The robust and high-performance control of fluid flows presents an engineering grand challenge, with the potential to enable advanced technologies in domains as diverse as transportation, energy, security, and medicine.  
The control of fluid flows is challenging due to the confluence of strong nonlinearity, high-dimensionality, and multi-scale physics (e.g., turbulence), thus typically leading to an intractable optimization problem.  
However, recent advances in machine learning (ML) are revolutionizing computational approaches for these traditionally intractable optimizations by providing principled approaches to feature extraction methods with improved optimization algorithms.  
%
We develop a deep learning model predictive control framework that leverages ML methods to achieve robust control performance in a complex fluid system without recourse to the governing equations, and with access to only a few physically realizable sensors.  
This sensor-based, data-driven learning architecture is critically important for practical implementation in control-based engineering applications.

Model predictive control (MPC)~\citep{garcia1989model,lee2011springer} is among the most versatile and widely used model-based control approaches, which involves an online optimization of the control strategy over a predictive receding horizon. 
Generally, improved models result in better control performance~\citep{WeisbergAndersen.1992}, although the online iterative optimization requires relatively inexpensive models~\citep{XI.2013,Mohanty.2009}.  
%
%
The challenge of MPC for controlling fluid flows is centered on the the high-dimensional nature of spatio-temporal flow fields.
Fortunately, these systems often exhibit dominant patterns of low-dimensional activity.  
Indeed, it is observed that flying insects, birds, and bats are able to harness these dominant patterns to execute exceptional control performance.  
Thus, there is a vibrant field in reduced-order models that balance accuracy and efficiency to capture essential physical mechanisms, while discarding distracting features.  
The MPC architecture, like much of ML, can be trained to exploit the dominant, low-dimensional patterns of dynamic activity in order to achieve significant gains in performance.

Among machine learning algorithms, deep learning~\citep{Krizhevsky2012nips,Lecun2015nature,Goodfellow-et-al-2016} has seen unprecedented success in a variety of modeling tasks across industrial, technological, and scientific disciplines.  
It is no surprise that deep learning has been rapidly integrated into several leading control architectures, including MPC and reinforcement learning. 
Deep reinforcement learning~\citep{Mnih.2015,Danilo.2014} has been widely used to learn games~\citep{Silver.2016,Mnih.2015}, and more recently for physical systems, for example to learn flight controllers~\citep{Kim2004nips,Tedrake2009isrr} or the collective motion of fish~\citep{Verma2018EfficientLearning}.  
DeepMPC~\citep{Verma2018EfficientLearning} has also emerged as a leading control strategy that combines the representational power of deep neural networks with the flexible optimization framework of MPC. 
There exist various approaches for using data-driven surrogate models for MPC
(e.g., based on the Koopman operator~\citep{KM18,KKB17,KKB18,NYK+18,PK19,Pei18}), and DeepMPC has considerable potential~\citep{BBK18,MJKW18}. 
\citep{MJKW18} recently demonstrated the ability of DeepMPC to control the flow past a laminar circular cylinder; the flow considered in this work is nearly linear, and may be well approximated using more standard linear modeling and control techniques, although this study provides an important proof of concept.

In this work, we extend DeepMPC for flow control in two key directions: (1) we apply this architecture to control significantly more complex flows that exhibit broadband phenomena; and (2) we develop our architecture to work with only a few physically realizable sensors, as opposed to earlier studies that involve the assumption of full flow field measurements. 
There is a significant gap between academic flow control examples and industrially relevant configurations.  
The present work takes a step towards complexity and importantly develops a data-driven, sensor-based architecture that is likely to scale to harder problems.  
Importantly, one rarely has access to the full flow field, and instead control must be performed with very few measurements~\citep{Manohar2017csm}. 
Biological systems, such as flying insects, provide proof by existence that it is possible to enact extremely robust control with limited flow measurements~\citep{Mohren2018pnas}.  
In this work, we design our learning approach to leverage time histories of limited sensors (i.e., measurable body forces), providing a more direct connection to engineering applications.
%
%
Finally, we provide a physical interpretation for the learned control strategy, which we connect to the underlying symmetries of the dynamical system.

\section{Model predictive control of complex systems}\label{sec:MPC_complexSystems}
\begin{figure}[h!]
    \center
    \includegraphics[width=.9\textwidth]{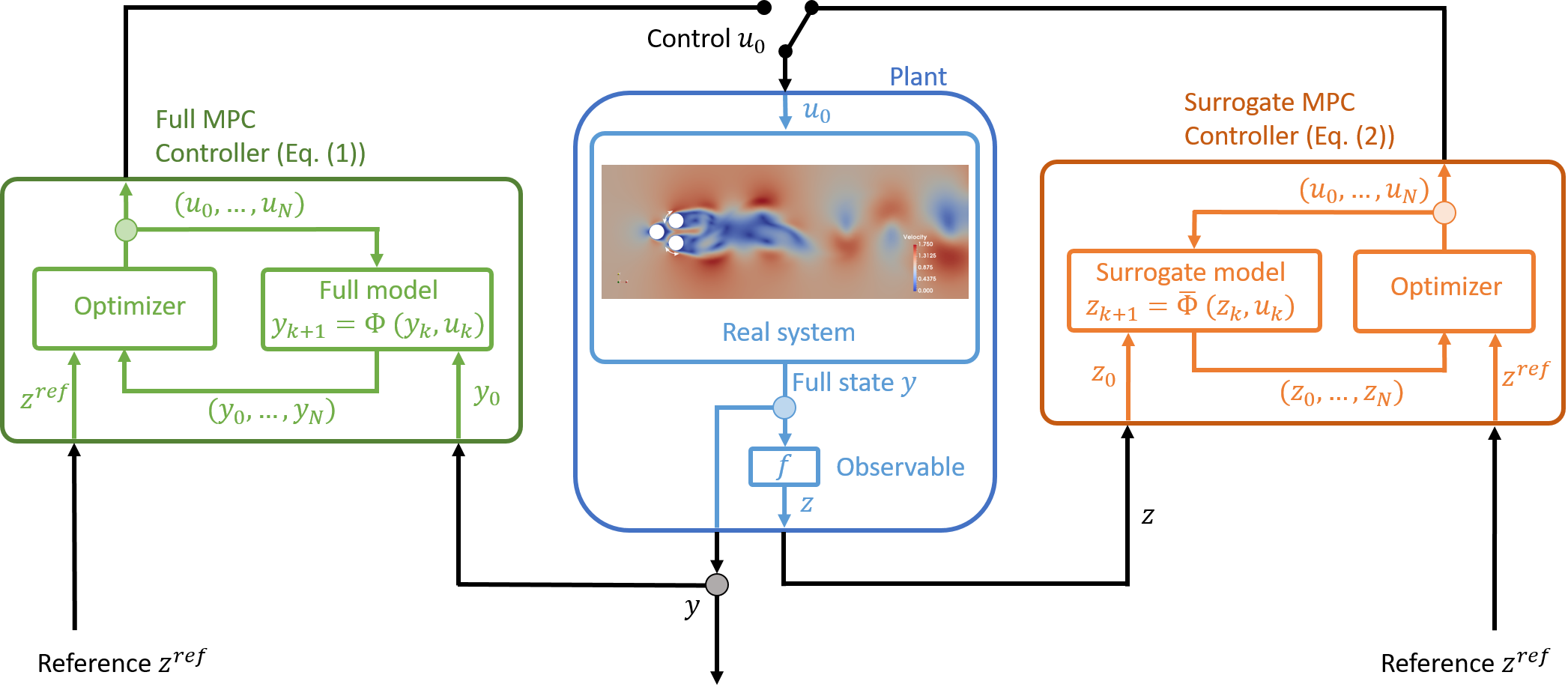}
    \caption{Structure of the control scheme, where a classical MPC controller based on a model for the full system state is shown in green and a controller using a surrogate model in orange.}
    \label{MPC_structure}
\end{figure}

Our main task is to control a complex nonlinear system in real-time. We do this by using the well-known MPC paradigm, in which an open-loop optimal control problem is solved in each time step using a model of the system dynamics:
\begin{equation} \label{eq:MPC}
    \begin{aligned}
    \min_{u\in \R^N} \sum_{i=0}^{N-1} \|f(y_{i+1}) - z_{i+1}^{\textnormal{ref}}\|_2^2 + \alpha | u_i |^2 + \beta | u_i - u_{i-1}|^2 \qquad s.t. \quad y_{i+1} = &\Phi(y_i, u_i).
    \end{aligned}
\end{equation}
Here, $f(y)=z$ is the observation of the time (and potentially space) dependent system state $y$ that has to follow a reference trajectory $z^{\textnormal{ref}}$, and $\alpha$ and $\beta$ are regularization parameters penalizing the control input as well as its variation. $\Phi$ is the time-$T$ map of the system dynamics, i.e., it describes how the system state evolves over one time step given the current state and control input. 
Problem \eqref{eq:MPC} is then solved repeatedly over a fixed \emph{prediction horizon} $N$ and the first entry is applied to the real system. 
As the initial condition in the next time step, the real system state is used such that a feedback behavior is achieved. Note that $u_{-1}$ is the control input that was applied to the system in the previous time step. The scheme is visualized in Fig.~\ref{MPC_structure}, where the MPC controller based on the full system dynamics is shown in green.

MPC has successfully been applied to a very large number of problems. However, a major challenge is the real-time requirement, i.e., \eqref{eq:MPC} has to be solved within the \emph{sample time} $\Delta t = t_{i+1} - t_i$. In order to achieve this, linearizations are often used. Since even these can be too expensive to solve for large systems, we will here use a surrogate model which does not model the entire system state but only the control relevant quantities. In a flow control problem, these can be the lift and drag coefficients of a wing, for instance. Such an approach has successfully been used in combination with surrogate models based on \emph{Dynamic Mode Decomposition} \citep{PK19} or \emph{Clustering} \citep{NYK+18}. We thus aim at directly approximating the dynamics $\overline{\Phi}$ for the observable $z=f(y)$ and replacing the constraint in Problem~\eqref{eq:MPC} by the surrogate model:
\begin{equation} \label{eq:MPCr}
    \begin{aligned}
    \min_{u\in \R^N} \sum_{i=0}^{N-1} \|z_{i+1} - z_{i+1}^{\textnormal{ref}}\|_2^2 + \alpha | u_i |^2 + \beta | u_i - u_{i-1}|^2 \qquad s.t. \quad z_{i+1} = \overline{\Phi}(z_i, u_i).
    \end{aligned}
\end{equation}
The resulting MPC controller is visualized in Fig.~\ref{MPC_structure} in orange.

\section{DeepMPC: model predictive control with a deep recurrent neural network}\label{sec:DeepMPC}

In order to solve \eqref{eq:MPCr}, the surrogate model $\overline{\Phi}$ for the control relevant system dynamics is required. For this purpose, we will use a deep RNN architecture which is implemented in TensorFlow \citep{TF15}. Once the model is trained and can predict the dynamics of $z$ (at least over the prediction horizon), the model can be incorporated in the MPC loop.

\subsection{Design of the RNN}
\begin{figure}[h]
    \begin{subfigure}{\textwidth}
       \center
       \begin{overpic}[width=.75\textwidth]{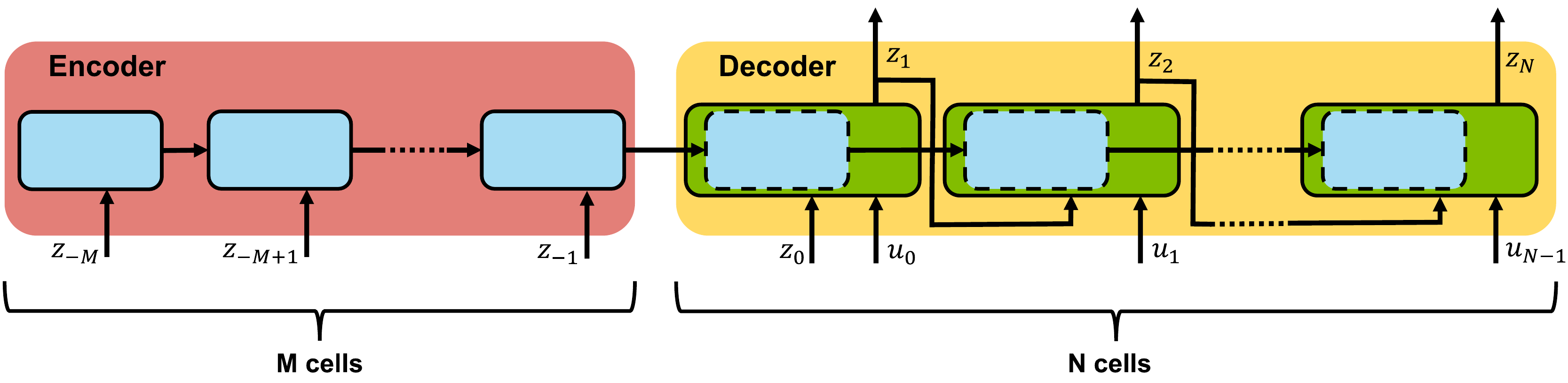}
        \put(-8,20){(a)}
        \end{overpic}
    \vspace*{0.3cm}
    \end{subfigure}

    \begin{subfigure}{\textwidth}
       \center
       \begin{overpic}[width=.7\textwidth]{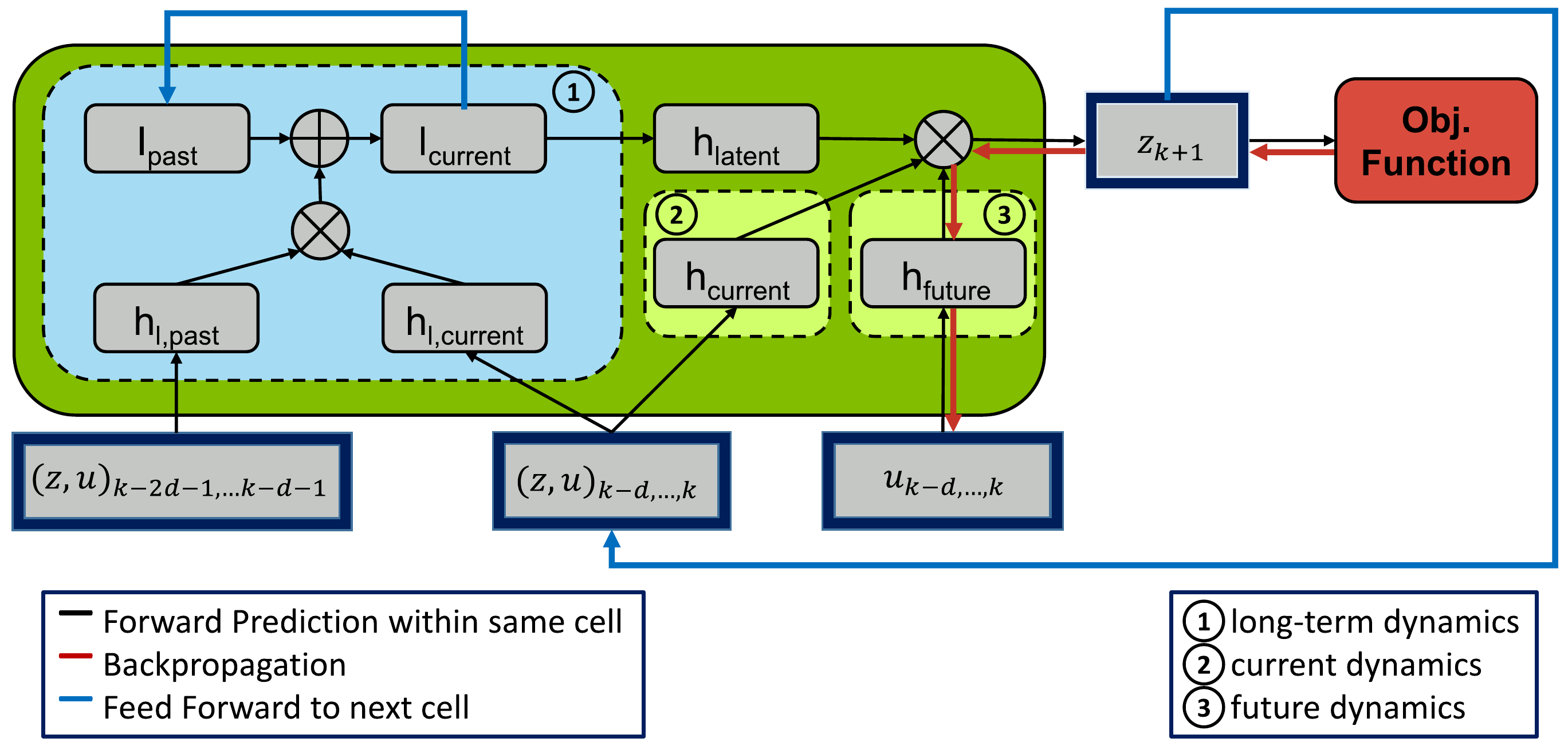}
        \put(-12,46){(b)}
        \end{overpic}
   \end{subfigure}
   \caption{(a) Unfolded RNN consisting of encoder and decoder. (b) Layout of a single RNN cell. An encoder cell only consists of the blue area. A decoder cell, on the other hand, contains the entire green cell.}\label{RNN_structure}
\end{figure}

As mentioned previously, the surrogate model is approximated using a deep neural network similar to \citep{BBK18}. The RNN consists of an encoder and a decoder (cf.~Fig.~\ref{RNN_structure}a), where the decoder performs the actual prediction task and consists of $N$ cells -- one for each time step in the prediction horizon. The encoder only predicts a latent state and this way takes the long term dynamics into account. Consequently, the encoder cells contain only the part of the decoder cells responsible for predicting the latent variable, cf.~Fig.~\ref{RNN_structure}b. 

In order to capture the system dynamics using few observations only, we use delay coordinates, an approach which has been successfully applied to many systems (\citep{BBP+17}). Consequently, each RNN cell takes as input a sequence of past observations as well as the current control input. The cells are divided into three functional parts capturing different parts of the dynamics, i.e., long term and current dynamics as well as the influence of the control inputs (see Fig.~\ref{RNN_structure}b). Therefore, the input of each cell $k$ is divided into a time series $(z,u)_{k-2d-1,\dots,k}$ of the observable and the corresponding control input and a second sequence of control inputs $u_{k-d,\dots,k}$, where $d$ is the number of delays.

The RNN based MPC problem \eqref{eq:MPCr} is solved using a gradient based optimization algorithm like gradient descent or BFGS. The required gradient information with respect to the control inputs can be calculated using standard back-propagation through time.
Since the RNN model needs at least temporal information from $M + 2d$ time steps ($M$ encoder cells and input sequence of length $2d$) to predict future states, there is an initialization phase in the MPC framework during which the control input is fixed to $0$.

\subsection{Training of the RNN}

\begin{figure}[b]
\begin{subfigure}{0.48\textwidth}
  \center
  \includegraphics[width=\linewidth]{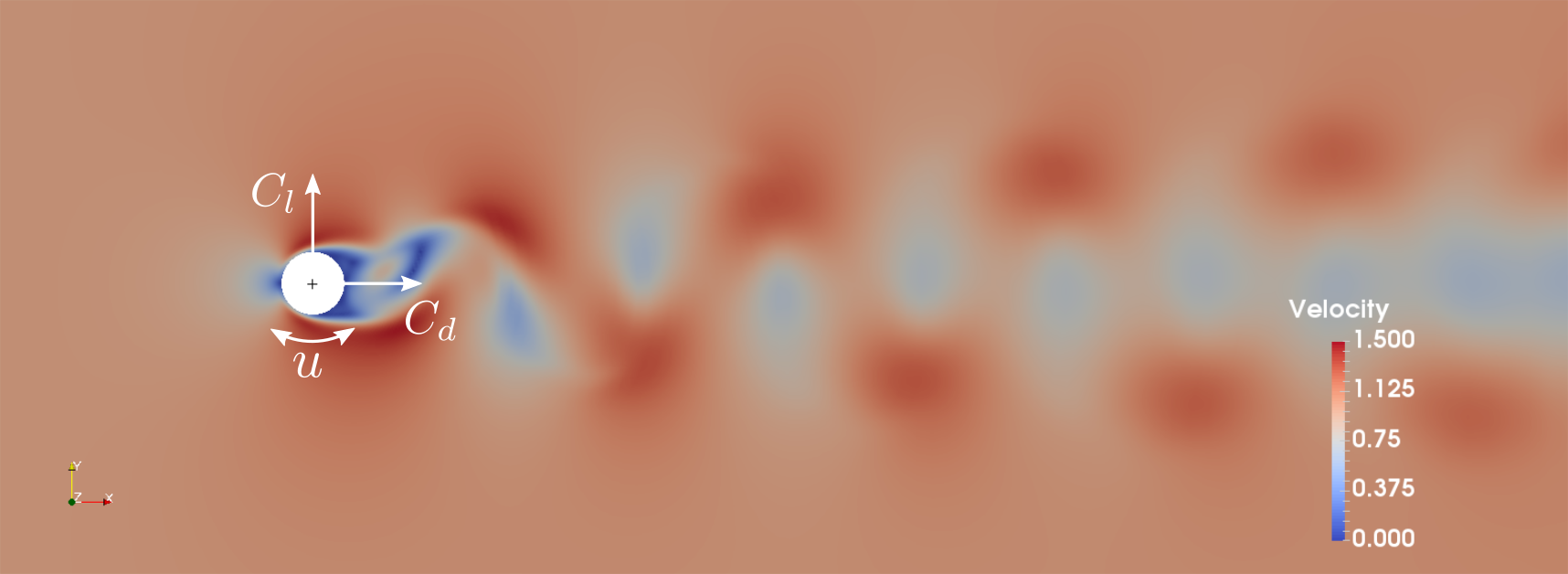}
  \caption{}\label{fig:karman_setting}
\end{subfigure}
\hfill
\begin{subfigure}{0.48\textwidth}
  \center
  \includegraphics[width=\linewidth]{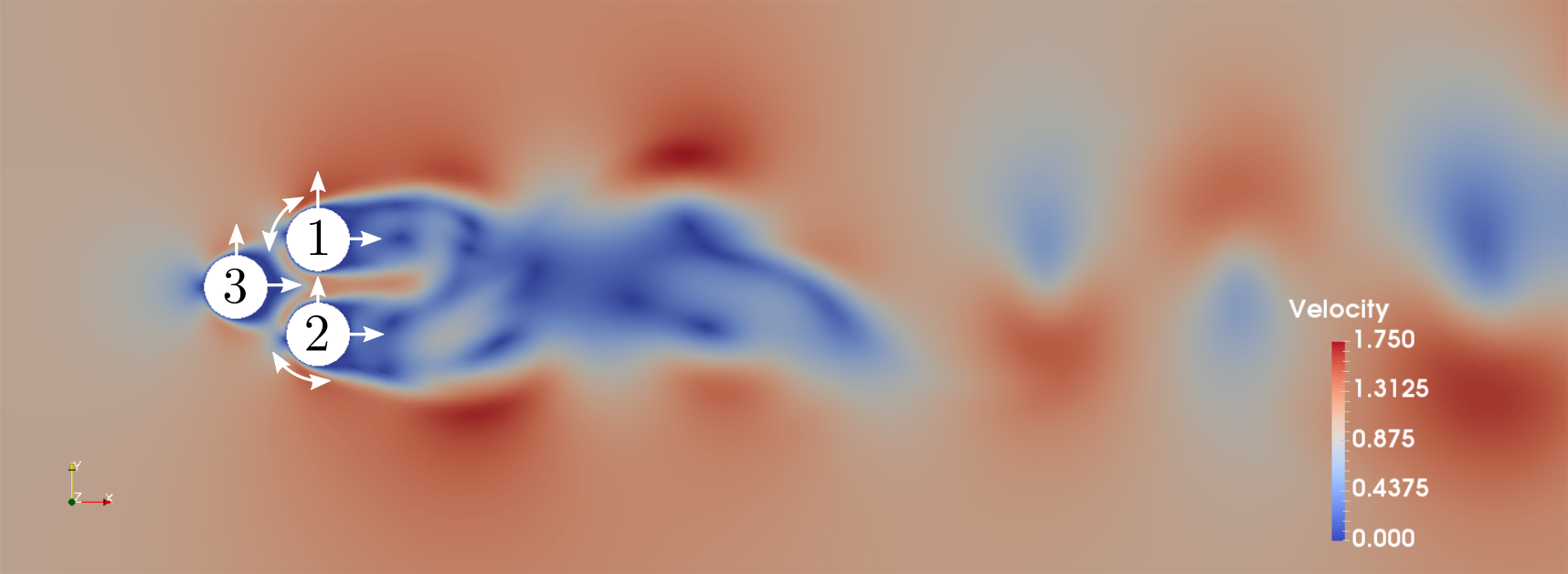}
  \caption{}\label{fig:pinball_setting}
\end{subfigure}
\caption{(a) Single cylinder setup. The system is controlled by setting the angular velocity $u$ of the cylinder. (b) Setup for the fluidic pinball, where the forces on all cylinders are observed. The system is controlled by rotating cylinders one and two with the respective angular velocities $u_1$ and $u_2$.}\label{fig:setting}
\end{figure}

The RNN is trained in an offline phase using time series data $\left((z_0,u_0),\ldots,(z_n,u_n)\right)$. For the data collection, the system is actuated with random yet continuously varying inputs.
In order to overcome difficulties with exploding and vanishing gradients as well as problems with the effect of nonlinearities when iterating from one time step to another, we use the three-stage approach for learning as was proposed in \citep{LKS15} and used in \citep{BBK18}. First, a Conditional Restricted Boltzmann Machine is used to compute good initial parameters for the RNN according to the work by \cite{THR07}. In a second step, only the prediction of a single time step is trained. In the final step, another training phase is performed, this time for the RNN with $N$ decoder cells that yield predictions for the system state over $N$ time steps.

\section{Results}\label{sec:Results}

In order to study the performance of the proposed MPC framework, four flow control problems of increasing complexity are considered. Instead of a real physical system, we here use a numerical simulation of the full model as our plant. In all four cases, the flow around one or multiple cylinders (cf.~Fig.~\ref{fig:setting}) is governed by the incompressible 2D Navier--Stokes equations with fluid entering from the left at a constant velocity $y_{in}$. The Reynolds number $Re = y_{in} / \nu D$ (based on the kinetic viscosity $\nu$ and the cylinder diameter $D$) ranges from 100 to 200, i.e., we are in the laminar regime. The full system is solved using a finite volume discretization and the open source solver OpenFOAM, cf.~\citep{JJT07}. The control relevant quantities are the lift and drag forces (i.e., the forces in $x_2$ and $x_1$ direction) acting on the cylinders. These consist of both friction and pressure forces which can be computed from the system state (or easily measured in the case of a real system).

\subsection{One Cylinder}

The first example is the flow around a single cylinder, cf.~Fig.~\ref{fig:karman_setting}, which was also studied in \citep{MJKW18}. At $Re=100$, the uncontrolled system possesses a periodic solution, the so called \emph{von Kármán vortex street}. On the cylinder, the fluid and the cylinder velocity are identical (no-slip condition) such that the flow can be steered by rotating the cylinder. The control relevant quantities are the forces acting on the cylinder -- the lift $C_l$ and drag $C_d$. We thus set $z = (C_l, C_d)$, and the aim is to control the cylinder such that the lift follows a given trajectory, e.g., a piece-wise constant lift.

In order to create training data, a time series of the lift and the drag is computed from a time series of the full system state with a random control sequence. To avoid high input frequencies, a random rotation between $-2$ and $2$ is chosen every $0.5 \,\text{sec}$. The intermediate control inputs are then computed using a spline interpolation on the grid of the time-$T$ map, where $\Delta t = 0.1 \,\text{sec}$. 
For the RNN training, a time series with $110\hspace*{0.05cm}000$ data points is used which corresponds to a duration of $11\hspace*{0.05cm}000 \,\text{sec}$. 

\begin{figure}[b]
\begin{subfigure}{0.5\textwidth}
  \center
  \includegraphics[width=\textwidth]{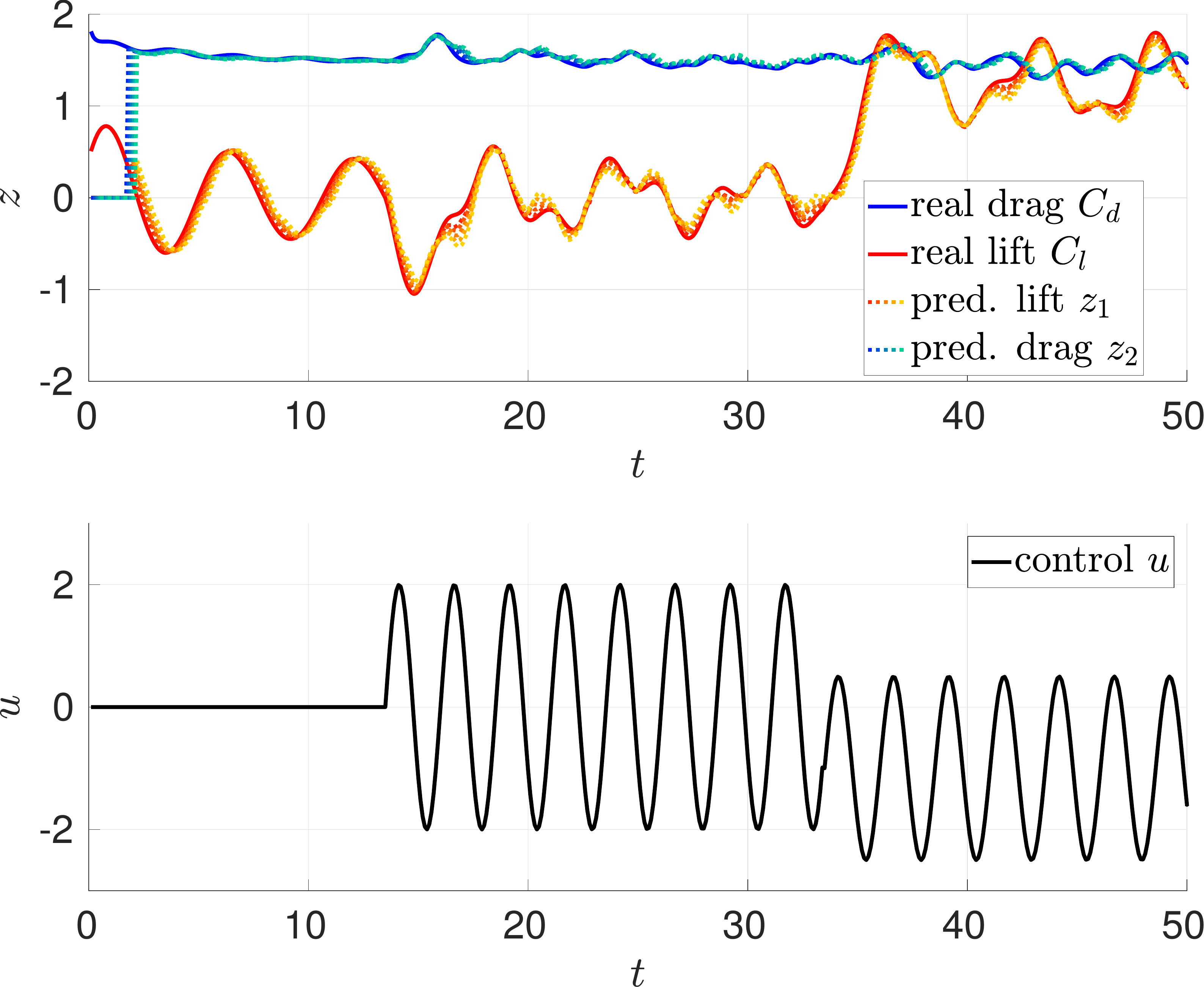}
  \caption{}\label{fig:karman_pred}
\end{subfigure}
\begin{subfigure}{0.5\textwidth}
    \center
    \includegraphics[width=\textwidth]{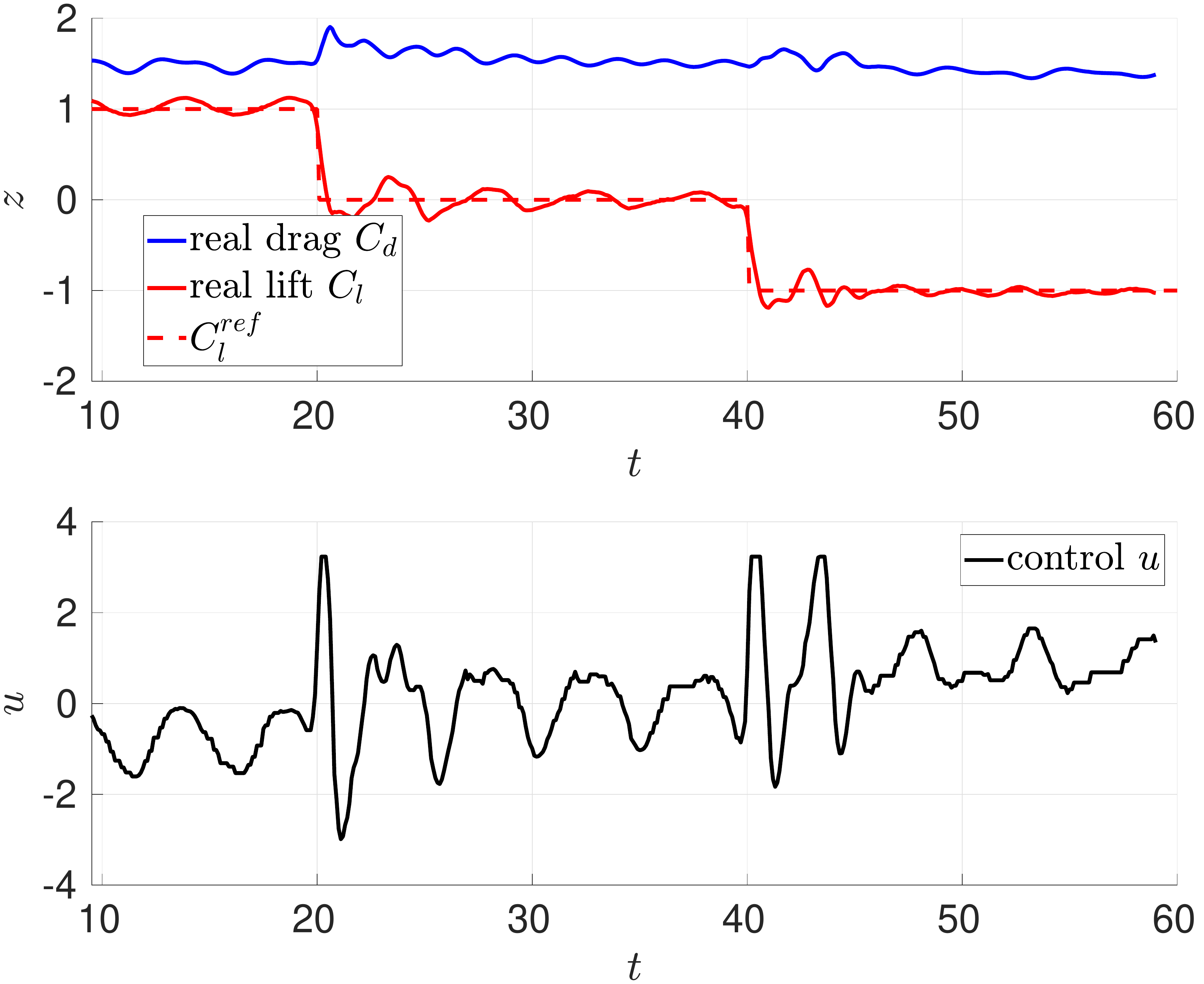}
   \caption{}\label{fig:karman_ctrl}
\end{subfigure}
\caption{System with one cylinder at $Re=100$. (a) Results for the prediction by the RNN for a given control input sequence. The prediction for the next $5$ time steps for lift and drag (at each $t_i$) are shown in brightening red or blue tones. (b) Results of the control task. The aim is to force the lift to $+1$, $0$ and $-1$ for $20 \,\text{sec}$, respectively.}
\end{figure}

In a first step, the quality of the RNN prediction is evaluated on the basis of an exemplary control input sequence. As one can see in Fig.~\ref{fig:karman_pred}, the prediction is very accurate over several time steps for many combinations of observations $z$ and control inputs $u$. There are only small regions where the predictions deviate stronger from the real lift and drag. 

The good prediction quality enables us to use the RNN in the MPC framework, where the aim is to force the lift to $+1$, $0$ and $-1$ for $20 \,\text{sec}$, respectively. This results in the following realization of \eqref{eq:MPCr}:
\begin{align}\label{eq:MPC_karman}
    \min_{u \in [-2,2]^5} \sum_{i=0}^{4} \norm{z_{1,i+1} - C_{l,i+1}^{\textnormal{ref}}}_2^2 + \beta |u_{i} - u_{i-1}|^2\qquad s.t. \quad z_{i+1} = \overline{\Phi}(z_i, u_i).
\end{align}
The parameter $\beta$ is set to $0.01$ in order to avoid too rapid variations of the input. Furthermore, the control is bounded by the minimum and maximum control input of the training data (i.e., $\pm2$). We solve the optimization problem \eqref{eq:MPC_karman} over a prediction horizon of length $N=5$, and we use a BFGS method for the optimization.
As shown in Fig.~\ref{fig:karman_ctrl}, the DeepMPC scheme shows very good performance. Due to the periodic fluctuation of the uncontrolled system, a periodic control is expected to suppress this behavior which is what we observe.


\subsection{Fluidic Pinball}

In the second example, we control the flow around three cylinders in a triangular arrangement, as shown in Fig.~\ref{fig:pinball_setting}. This configuration is known as the \emph{fluidic pinball}; see \citep{DPMN18} for details. 
The control task is to make the lift of the three cylinders ($C_{l,1}$, $C_{l,2}$ and $C_{l,3}$) follow given trajectories by rotating the rear cylinders while the cylinder in the front is fixed. We thus want to approximate the system dynamics of the forces acting on all three cylinders, i.e., $z=(C_{l,1},C_{l,2},C_{l,3},C_{d,1},C_{d,2},C_{d,3})$. Similar to the single cylinder case, the system possesses a periodic solution at $Re = 100$. When increasing the Reynolds number, the system dynamics become chaotic (cf.~\citep{DPMN18}) and the control task is much more challenging. We thus additionally study the chaotic cases $Re=140$ and $Re = 200$. 
As we now have two inputs and three reference trajectories, we obtain the following realization of problem \eqref{eq:MPCr}:
\begin{align*}
    \min_{u_1,u_2 \in [-2,2]^N} \sum_{i=0}^{N-1} \left(\sum_{j=1}^{3} \norm{z_{j,i+1} - C_{l,j,i+1}^{\textnormal{ref}}}_2^2 + \beta \sum_{j=1}^{2} |u_{j,i} - u_{j,i-1}|^2\right)\quad s.t. \quad z_{i+1} = \overline{\Phi}(z_i, u_i),
\end{align*}
where the value of $\beta$ is set to $0.1$. The prediction horizon is $N=5$ for $Re=100$ and $N=10$ for $Re=140$ and $Re=200$, respectively.

\begin{figure}[h!]
 \captionsetup{format=myformat}
    \begin{subfigure}{\textwidth}
        \includegraphics[width=\textwidth]{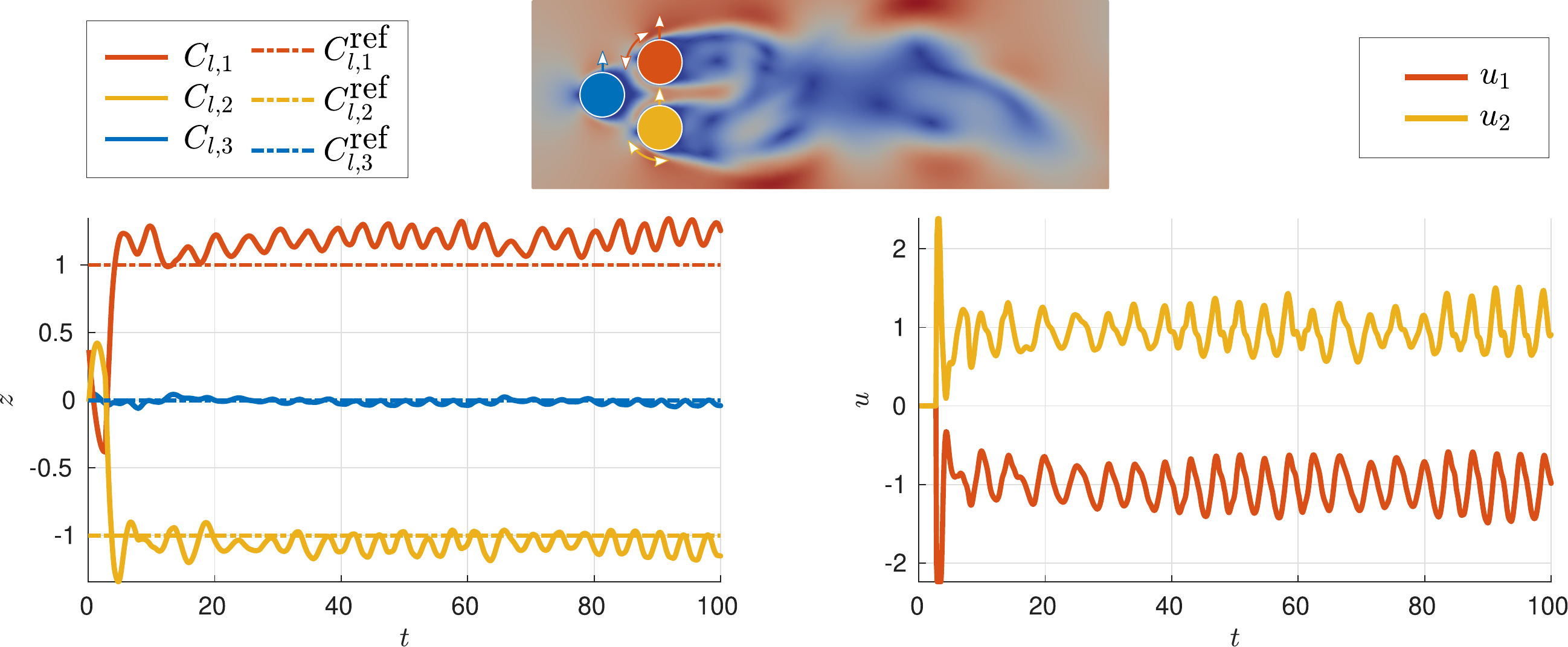}
        \caption{$Re=100$: $e_{mean} = \hspace*{0.13cm} \frac{\Delta t}{T} \hspace*{0.08cm} \sum_{i=\frac{4}{\Delta t}}^{\frac{T}{\Delta t}} \hspace*{0.13cm} \frac{1}{3}\sum_{j=1}^{3} \norm{C_{l,j,i} - C_{l,j,i}^{\textnormal{ref}}}_2^2 = 0.016235$, with $T=100$ \\ \hspace*{0.25cm}$ e_{max} = \max_{\frac{4}{\Delta t} \le i \le \frac{T}{\Delta t}} \frac{1}{3}\sum_{j=1}^{3} \norm{C_{l,j,i} - C_{l,j,i}^{\textnormal{ref}}}_2^2 = 0.050267$}\label{fig:pinball_100}
    \end{subfigure}
\ \\
    \begin{subfigure}{\textwidth}
        \includegraphics[width=\textwidth]{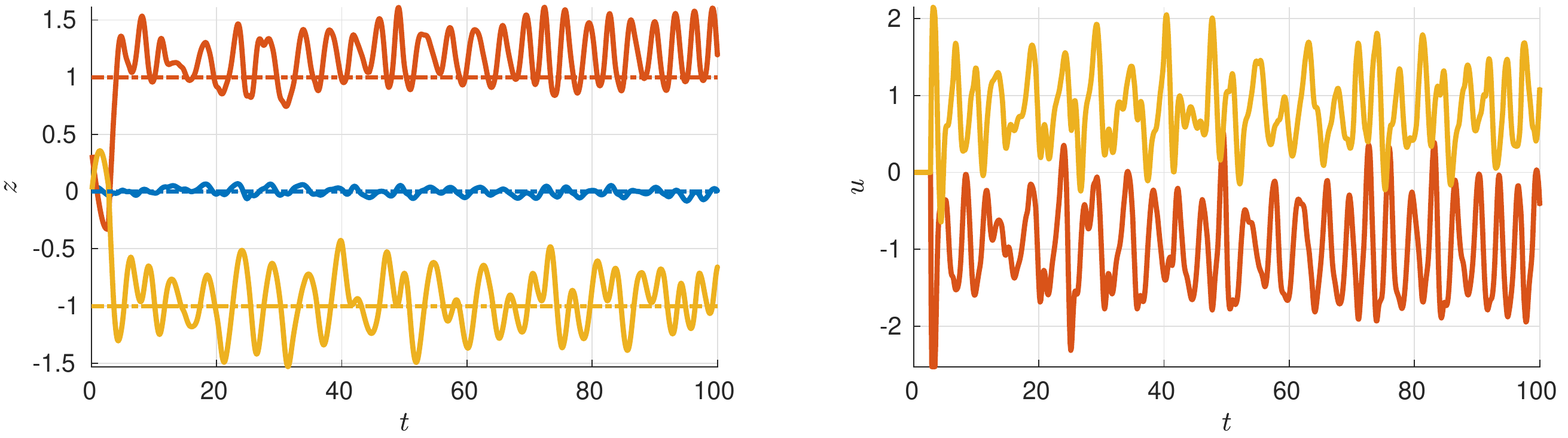}
        \caption{$Re=140$: $e_{mean} = 0.043695$, $e_{max} = 0.14484$}\label{fig:pinball_140}
   \end{subfigure}
    \begin{subfigure}{0.5\textwidth}
        \includegraphics[width=\textwidth]{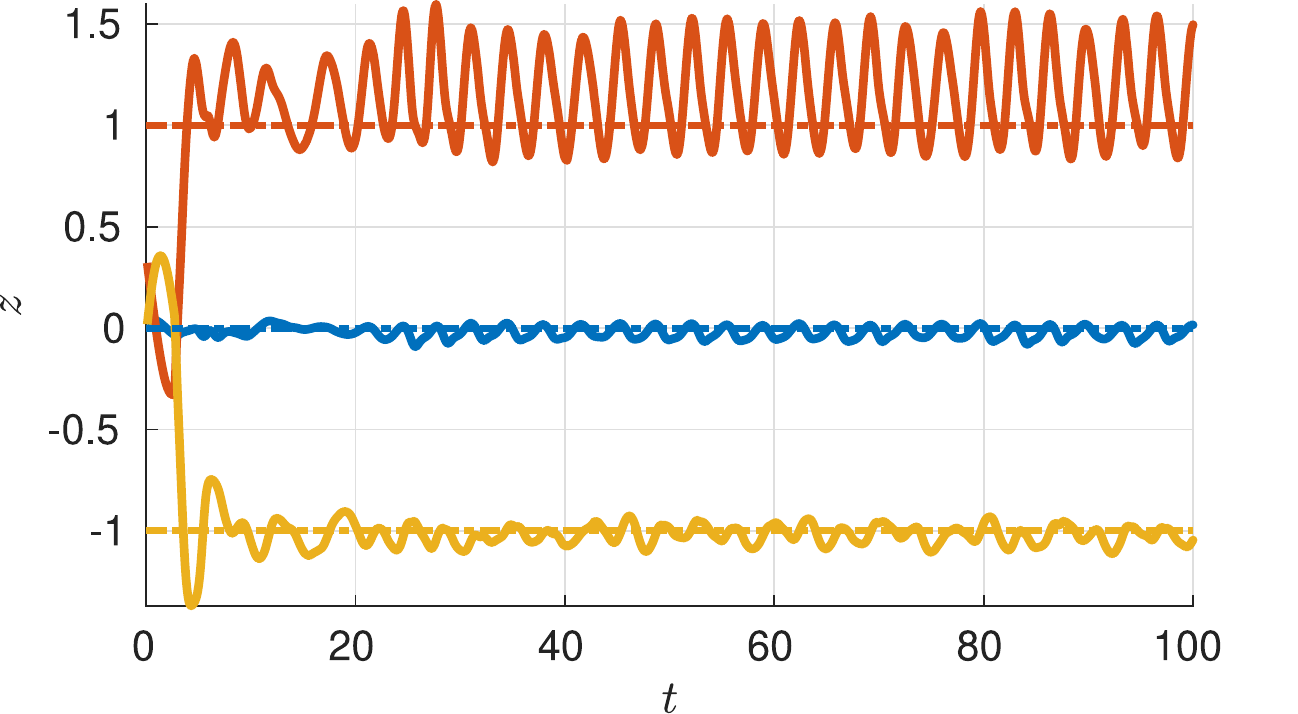}
      \caption{$Re=140$ with symmetrized data:\newline $e_{mean} = 0.025452$, $e_{max} = 0.11917$}\label{fig:pinball_140_sym}
    \end{subfigure}
    \begin{subfigure}{0.5\textwidth}
        \includegraphics[width=\textwidth]{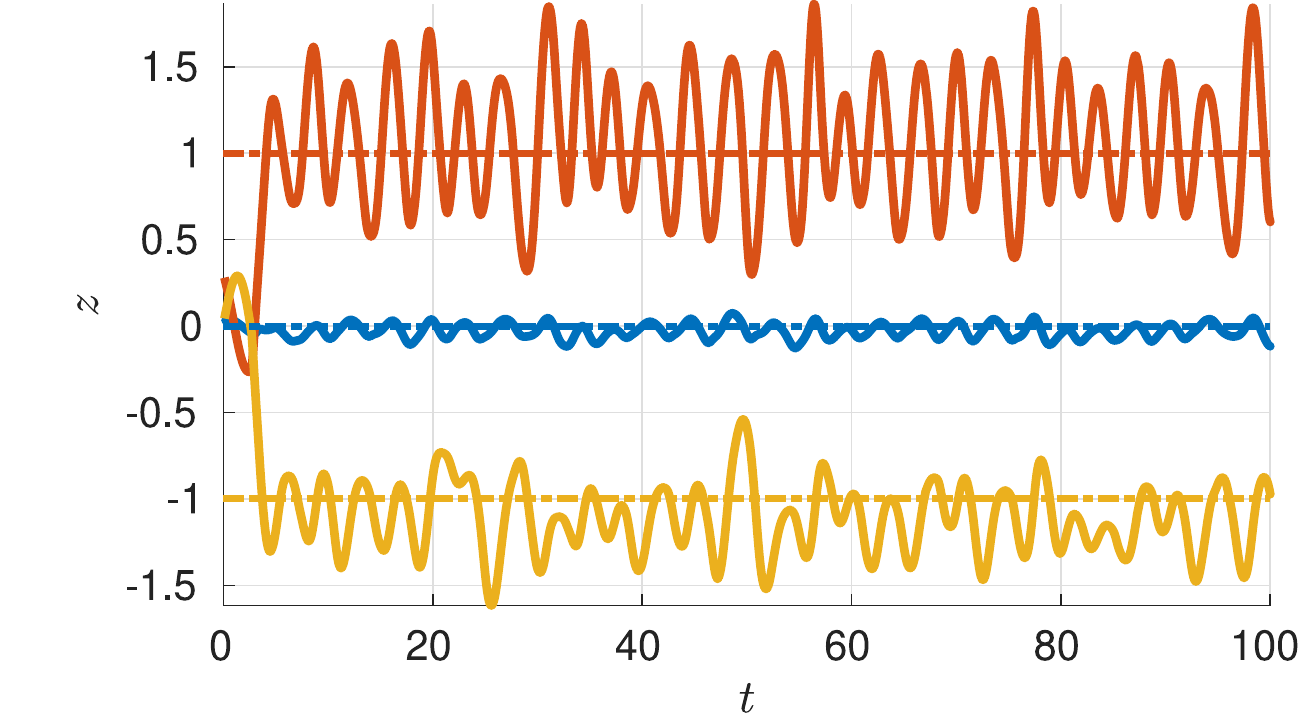}
        \caption{$Re=200$ with symmetrized data:\newline $e_{mean} = 0.063857$, $e_{max} = 0.25680$}\label{fig:pinball_200_sym}
    \end{subfigure}
   \caption{DeepMPC reference tracking for varying $Re$ and data sets.}
\end{figure}

For all three Reynolds numbers, the training data is computed by simulating the system with random yet smoothly varying control inputs as before, i.e., random values between $-2$ and $2$ for each cylinder every $0.5 \,\text{sec}$. Due to the significantly smaller time step of the finite volume solver for the fluidic pinball, the control is interpolated on a finer grid with step size $0.005\, \text{sec}$. Since the control input has to be fixed over one lag time due to the discrete-time mapping via the RNN, the mean over one lag time (i.e., over 20 data points) is taken for $u$. Time series with $150\hspace*{0.05cm}000$, $200\hspace*{0.05cm}000$ and $800\hspace*{0.05cm}000$ data points are used for $Re=100$, $Re=140$ and $Re=200$, respectively.

At $Re = 100$, where the dynamics are quasi-periodic, the control is quite effective, almost comparable to the single cylinder case, cf.~Fig.~\ref{fig:pinball_100}. 
In comparison, the error $e_{mean}$ for the mildly chaotic case $Re=140$ (Fig.~\ref{fig:pinball_140}) is approximately one order of magnitude larger. The reference is still tracked, but larger deviations are observed. However, since the system is chaotic, this is to be expected. It is more difficult to obtain an accurate prediction and -- more importantly (cf.~\citep{Pei18}) -- the system is more difficult to control.
In order to improve the controller performance, we incorporate system knowledge, i.e., we exploit the symmetry along the horizontal axis. 
Numerical simulations suggest that this symmetry results in two attracting regions in the observation space and that the system changes only occasionally from one region to the other, analogous to the Lorenz attractor \citep{BBP+17}. 
Therefore, we symmetrize (and double) the training data as follows:
\begin{equation}\label{eq:symmetry}
\begin{aligned}
    \hat{u} = \left( -u_2,  -u_1 \right)^\top, \quad
    \hat{C_l} = \left( -C_{l,2}, -C_{l,1}, -C_{l,3}\right)^\top, \quad
    \hat{C_d} = \left( C_{d,2}, C_{d,1},  C_{d,3}\right)^\top.
\end{aligned}
\end{equation}
This step is not necessary at $Re=100$, since the collected data is already nearly symmetric. Nevertheless, the amount of training data can be doubled by exploiting the symmetry and therefore, the simulation time to generate the training data can be reduced.

In Fig.~\ref{fig:pinball_140_sym}, the results for $Re=140$ with symmetric training data is shown. In this example, the tracking error is reduced by nearly $50\%$. In particular, the second lift is well-controlled. This indicates that it is advisable to incorporate known physical features such as symmetries in the data, assimilation process. However, we still observe that the existence of two attracting regions results in a better control performance for one of the cylinders, depending on the initial condition.

For the final example, the Reynolds number is increased to $Re = 200$ in order to further increase the complexity of the dynamics, and symmetric data is used again. Due to the higher Reynolds number, switching between the two attracting regions occurs much more frequently, and the use of symmetric data yields less improvement. The results are presented in Fig.~\ref{fig:pinball_200_sym}, and we see that even though tracking is achieved, the oscillations around the desired state are larger.  

\begin{figure}[h!]
\begin{subfigure}{0.48\textwidth}
        \center
        \begin{overpic}[width=\textwidth]{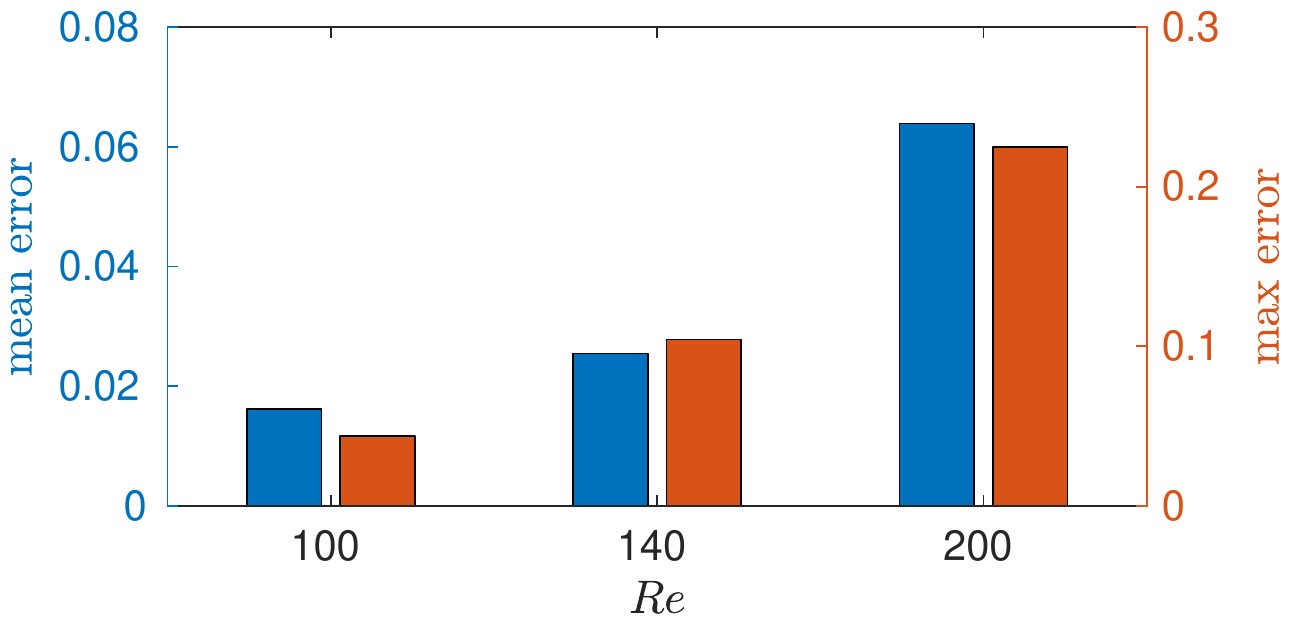}
        \put(15,40){(a)}
        \end{overpic}
    \end{subfigure}
\hfill
    \begin{subfigure}{.48\textwidth}
      \center
      \begin{overpic}[width=\textwidth]{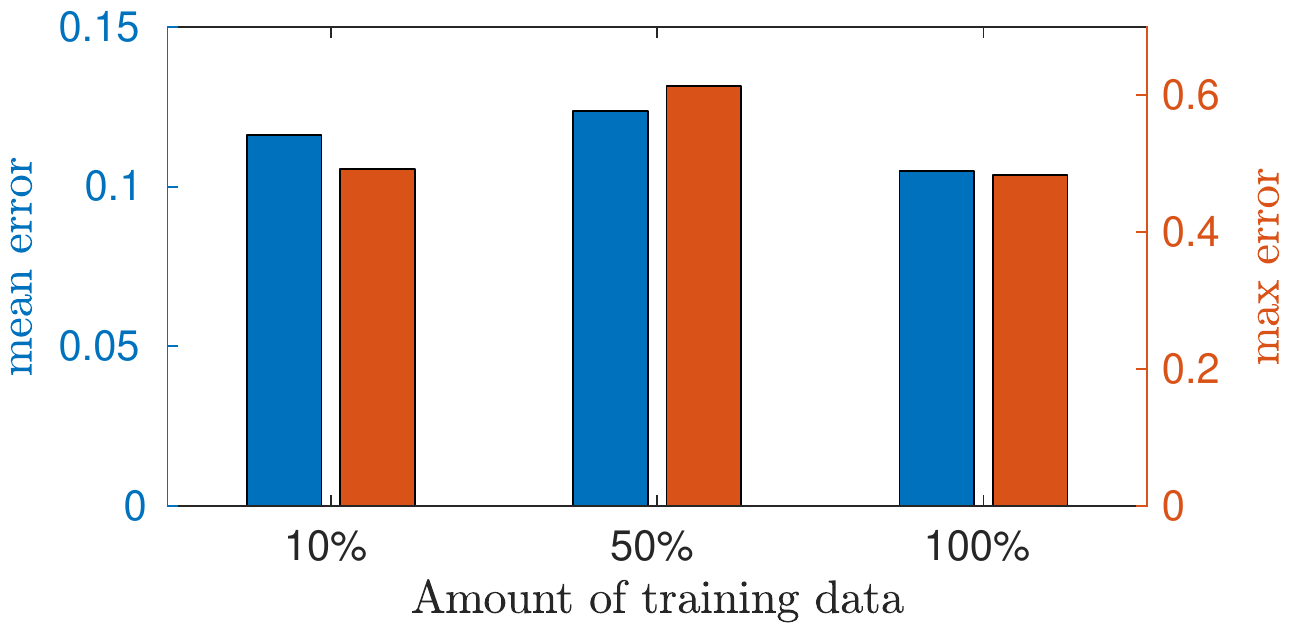}
              \put(15,40){(b)}
   \end{overpic}
    \end{subfigure}
    \caption{(a) Mean (blue) and maximal (red) error for various Reynolds numbers with full data. \newline (b) Mean and maximal error for different training data set sizes, both averaged over $5$ training runs ($Re = 200$).}\label{fig:hist}
\end{figure}

In Fig.~\ref{fig:hist}a the mean and the maximal error for the three Reynolds numbers are shown. Since the system dynamics become more complex with increasing Reynolds number, both the prediction by the RNN and the control task itself become more difficult and the tracking error increases. 

In order to study the robustness of the training process as well as the influence of the amount of training data on the tracking error, $5$ identical experiments for $Re=200$ have been performed for different amounts of training data ($10\%$, $50\%$ and $100\%$ of the symmetrized data points), respectively, see Fig.~\ref{fig:hist}b.
We observe no trend with respect to the amount of training data, in particular considering that the standard deviation is approximately $0.03$ for the average and $0.15$ for the maximal error. Short time series already cover large parts of the dynamics and are thus sufficient to train the model. In order to further improve performance, the size of the RNN as well as the length of the training process would have to be increased significantly and also, significantly smaller lag times would be required.

\begin{figure}[t]
    \centering
    \includegraphics[width=\textwidth]{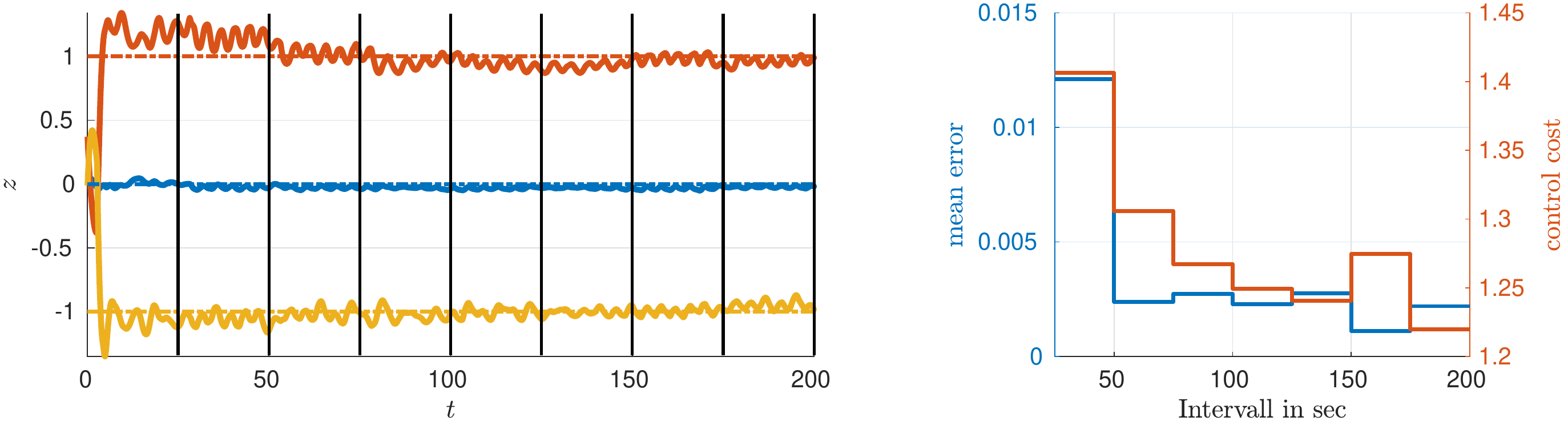}
     \caption{$Re=100$ with online learning. The RNN is updated every $25\,\text{sec}$ (denoted by black lines on the left). On the right the mean error (blue) and the control cost (red) over each interval are shown.}
     \label{fig:pinball_100_online}
\end{figure}

Since we want to avoid this further increase in computational effort and data collection, we instead use small amounts of data sampled in the relevant parts of the observation space, i.e., close to the desired state. To this end, we perform online updates using the incoming sensor data. In our final experiment, we study how the control performance can be improved by performing online updates of the RNN using the incoming data. In the feedback loop, a new data point is collected from the real system at each time step, and our strategy is to collect new data over $25 \,\text{sec}$ for each update. By exploiting the symmetry as proposed in \eqref{eq:symmetry}, we obtain $500$ points within each interval that are used for further training of the RNN. In the right plot of Fig.~\ref{fig:pinball_100_online}, we compare the tracking error over several intervals, and we see that the error can be decreased very efficiently in a short time by using online learning (see also Fig.~\ref{fig:pinball_100} for a comparison). Besides reducing the tracking error, the control cost $\norm{u}_2$ decreases, which further demonstrates the importance of using the correct training data. Significant improvements of both the tracking performance as well as the controller efficiency are obtained very quickly with comparably few measurements.
\section{Conclusion and Further Work}
We present a deep learning MPC framework for feedback control of complex systems.  
Our proposed sensor-based, data-driven learning architecture achieves robust control performance in a complex fluid system without recourse to the governing equations, and with access to only a few physically realizable sensors.  
In order to handle the real-time constraints, a surrogate model is built exclusively for control relevant and easily accessible quantities (i.e., sensor data). This way, the dimension of the RNN-based surrogate model is several orders of magnitude smaller compared to a model of the full system state. 
On the one hand, this enables applicability in a realistic setting since we do not rely on knowledge of the entire state. It further allows us to address systems of higher complexity, i.e., it is a sensor-based and scalable architecture.
The approach shows very good performance for high-dimensional systems of varying complexity, including chaotic behavior. To avoid prohibitively large training data sets and long training phases, an online update strategy using sensor data is applied. This way, excellent performance can be achieved for $Re=100$.
For future work, it will be important to further improve and robustify the online updating process, in particular for chaotic systems.
Furthermore, it is of great interest to further decrease the training data requirements by designing RNN structures specifically tailored to control problems. 
The deep learning MPC is a critically important architecture for real-world engineering applications where only limited sensors are available to enact control authority.

\subsubsection*{Acknowledgments}
This work is supported by the Priority Programme SPP 1881 ``Turbulent Superstructures'' of the Deutsche Forschungsgemeinschaft. The calculations were performed on resources provided by the Paderborn Center for Parallel Computing (PC$^2$). 

\medskip
\small
\bibliographystyle{plainnat}
\bibliography{sample}

\end{document}